\documentclass[letterpaper,10pt,conference]{ieeeconf}

\IEEEoverridecommandlockouts

\usepackage{graphicx}
\usepackage{amsmath}
\usepackage{amssymb}
\usepackage[table]{xcolor}
\usepackage{caption}
\usepackage{url}
\usepackage{booktabs}
\usepackage{multirow}
\renewcommand{\thetable}{\arabic{table}}
\definecolor{rblue}{rgb}{0,0.5,1}
\definecolor{awesome}{rgb}{1.0, 0.13, 0.32}
\definecolor{hollywoodcerise}{rgb}{0.96, 0.0, 0.63}
\definecolor{lasallegreen}{rgb}{0.03, 0.47, 0.19}
\definecolor{hanpurple}{rgb}{0.32, 0.09, 0.98}
\definecolor{green(pigment)}{rgb}{0.0, 0.65, 0.31}

\makeatletter
\let\NAT@parse\undefined
\makeatother
\usepackage[pagebackref=false, breaklinks=true, colorlinks, bookmarks=false]{
        hyperref
}
\hypersetup{
        colorlinks=true,
        linkcolor={red},
        citecolor={hanpurple},
        urlcolor={magenta}
}

\title{\LARGE \bf FUSEye: Training-Light Fisheye Detection with Overlapping Views and Zero-Initialized Adapters}

\author{Wenya Su$^{1,*}$, Kai Luo$^{1,*}$, Di Wen$^{2}$, Ruiping Liu$^{2}$, Yufan Chen$^{2}$, Junwei Zheng$^{2}$, Kunyu Peng$^{2,\dag}$, and Kailun Yang$^{1,\dag}$
\thanks{This work was supported in part by the National Natural Science Foundation of China (Grant No. 62473139 and No. 62388101), in part by the Hunan Provincial Research and Development Project (Grant No. 2025QK3019), in part by the State Key Laboratory of Autonomous Intelligent Unmanned Systems (the opening project number ZZKF2025-2-10), and in part by the Deutsche Forschungsgemeinschaft (DFG, German Research Foundation) – SFB-1574 – 471687386.}
\thanks{$^{1}$The authors are with the School of Artificial Intelligence and Robotics and the National Engineering Research Center of Robot Visual Perception and Control Technology, Hunan University, China (email: kailun.yang@hnu.edu.cn).}%
\thanks{$^{2}$The authors are with the Institute for Anthropomatics and Robotics, Karlsruhe Institute of Technology, Germany (email: kunyu.peng@kit.edu).}
\thanks{$^{*}$Equal contribution.}
\thanks{$^{\dag}$Corresponding authors: Kailun Yang and Kunyu Peng.}
}

\begin{document}

\let\oldtwocolumn\twocolumn
\renewcommand\twocolumn[1][]{%
  \oldtwocolumn[{#1}{
  \begin{center}
    \vskip -3ex
    \centering
    \includegraphics[width=0.99\textwidth]{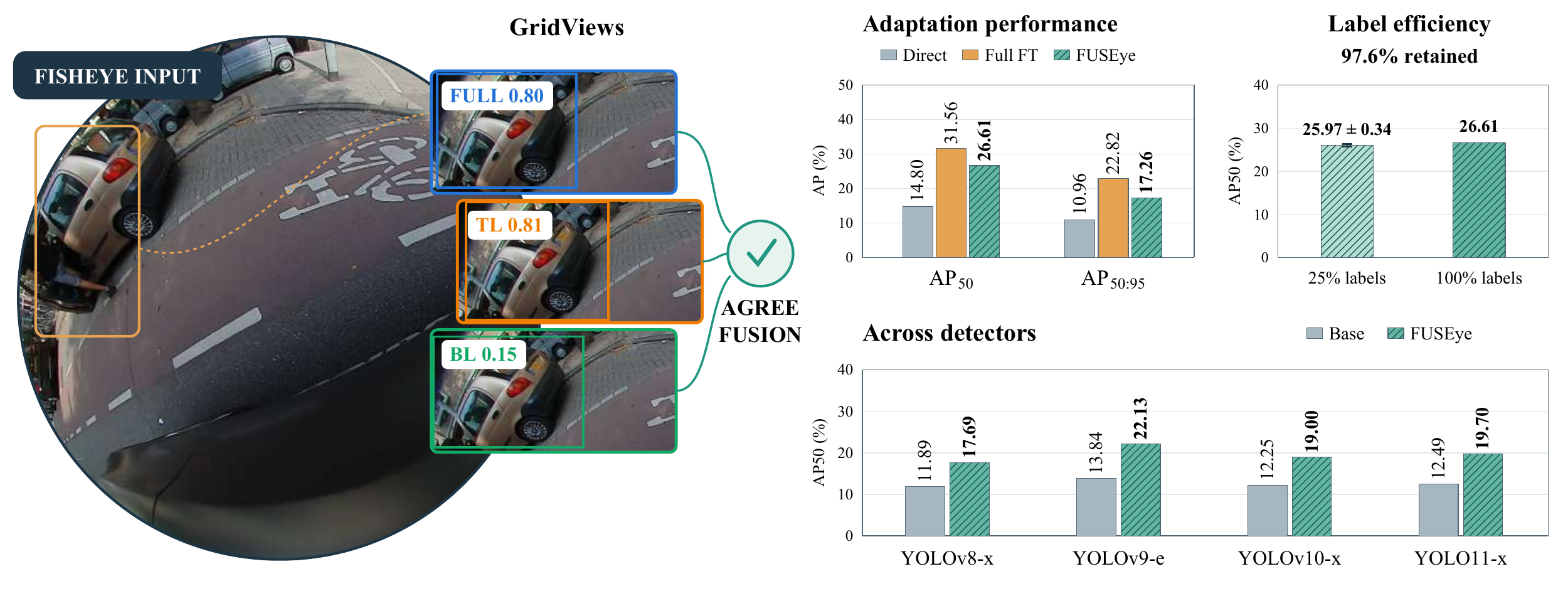}
    \vskip -1ex
    \captionof{figure}{
      Motivation and main results of FUSEye. The full fisheye image and its overlapping local views provide complementary detection hypotheses for the same highly distorted boundary object, which are consolidated by AgreeFusion. 
      With the COCO-pretrained backbone frozen, FUSEye improves overall mAP50 on WoodScape MVR from $0.148$ to $0.266$, approaching the $0.315$ obtained by full fine-tuning. 
      The class-wise results show clear improvements for \emph{person} and \emph{car}.
    }
    \label{fig:teaser}
  \end{center}
  }]
}

\maketitle
\raggedbottom
\begin{abstract}
Fisheye cameras give mobile robots a single-sensor, low-cost view of their surroundings, yet the COCO-pretrained detectors that practitioners routinely reuse fail on them: strong radial distortion warps local image structure, while boundary compression shrinks objects to near-invisible sizes. Full fine-tuning closes much of the gap but requires abundant fisheye labels and compute. We present FUSEye, a training-light framework that turns a frozen-backbone COCO-pretrained extra-large YOLO26 detector (YOLO26-x) into a fisheye detector. FUSEye adds roughly $227k$ new parameters while updating the inserted modules and the pretrained detection head. It addresses the transfer gap at three causally linked levels. At the input level, overlapping grid view generation and box remapping (GridViews) enlarge compressed boundary regions. At the feature level, zero-initialized residual adapters (Z-Adapters) correct distortion-induced feature misalignment. At the decision level, learned cross-projection agreement fusion (AgreeFusion) promotes low-confidence detections only when they are supported by consistent evidence across multiple views. On the WoodScape surround-view fisheye benchmark, FUSEye raises YOLO26-x from $0.148$ to $0.266$ mAP50 and retains $84.3\%$ fully fine-tuned accuracy. Moreover, randomly using only $25\%$ of the labeled training images, FUSEye achieves $0.2597{\pm}0.0034$ mAP50, retaining $97.6\%$ of its full-label performance. FUSEye also consistently improves YOLOv8-11 detectors, showing that the recipe is architecture-agnostic. Source code will be available at \url{https://github.com/Su-wenya/FUSEye}.
\end{abstract}

\section{Introduction}
Fisheye cameras offer mobile robots a cost-effective wide-angle sensing solution: a single sensor captures a broad field of view, making fisheye cameras attractive for robot navigation and obstacle avoidance~\cite{woodscape2019,fisheyesurvey2023}. 
Object detection is a fundamental perception task in these applications~\cite{rashed2021generalized}. 
Annotating fisheye datasets is labor-intensive, and prior work has adopted COCO-pretrained YOLO-family detectors as a starting point for adaptation~\cite{fisheye8k2023}.
However, strong radial distortion causes spatially varying changes in object appearance and compresses objects near the image boundary, making features learned from perspective images less transferable to fisheye views~\cite{woodscapechallenge2022,fisheyedet2020}.
This limitation is particularly critical for mobile robots: pedestrians and obstacles entering the field of view may initially appear as small, strongly compressed boundary objects. Missing these early observations reduces the time available for braking or replanning.

This projection-induced domain shift poses a key challenge when adapting pretrained object detectors to fisheye imagery. 
As illustrated in Figure~\ref{fig:teaser}, directly applying the COCO-pretrained extra-large YOLO26 model (YOLO26-x) to the mirror-view right (MVR) validation split of WoodScape~\cite{woodscape2019} yields only $0.1480$ $\mathrm{mAP}_{50}$ for \emph{car}, \emph{person}, and \emph{bus}, while the complete cross-view fine-tuning reaches $0.3156$.
Beyond boundary compression, training-light transfer involves two additional challenges. 
Distorted object appearances can misalign the backbone features learned from perspective images, while overlapping crops used to recover peripheral objects can introduce duplicate and conflicting detection hypotheses.

Existing distortion-aware approaches involve different adaptation costs. Projection-aware kernel adaptation can require camera calibration~\cite{convkerneladapt2024}; rectification networks introduce an additional preprocessing model~\cite{fishformer2022}; and deformable convolution methods modify the feature extractor and typically require full-model retraining~\cite{adaptdeform2021}.
Generic parameter-efficient adapters reduce the number of updated parameters, but are not designed for the spatially non-uniform distortion of fisheye images~\cite{pettvision2024}. 
Slicing and multi-view inference enlarge small boundary objects, but conventional suppression or fusion treats redundant detections without explicitly considering whether they are supported across multiple projections~\cite{sahi2022,softnms2017,wbf2021}. 
Although these methods provide useful building blocks, their integration within a frozen-backbone fisheye adaptation framework remains less explored.
This observation motivates the central question of this work: can a COCO-pretrained detector be adapted to fisheye imagery without full-model fine-tuning, while jointly addressing input-level boundary compression, feature-level misalignment, and decision-level cross-view conflicts under a limited labeling and training budget?

We answer this question with FUSEye, a training-light framework, which views the transfer gap as a chain of three causally linked problems rather than a single capacity deficit. 
At the input level, boundary objects are too compressed for the network input scale, so overlapping grid view generation and box remapping (GridViews) generate the full frame plus four overlapping patches and remap all boxes back to the original coordinates, requiring no calibration or dewarping. 
At the feature level, radial distortion misaligns the features extracted by the shared backbone, so zero-initialized residual adapters (Z-Adapters) are inserted at three high-level backbone stages; the zero-initialized output projection keeps the pretrained model unchanged at the first step and then learns a residual correction for the fisheye domain with only $222{,}480$ parameters on YOLO26-x. 
At the decision level, multi-view inference produces duplicate and conflicting detections, so cross-projection agreement fusion (AgreeFusion) clusters per-view detections by class and intersection over union (IoU), re-scores multi-view clusters with a learned scorer of about $4{,}400$ parameters, applies confidence-weighted box fusion, and preserves the baseline precision policy for single-view detections. GridViews supplies candidates for AgreeFusion, while Z-Adapters preserve initial per-view predictions. Each component can be ablated independently with the backbone frozen.

On the WoodScape MVR validation split, FUSEye improves YOLO26-x from $0.1480$ to $0.2661$ mAP50, retains $84.3\%$ of fully fine-tuned accuracy while training far fewer parameters, and preserves $97.6\%$ of its full-label performance when trained with only $25\%$ of the labels.
It also generalizes across YOLO families, improving mAP50 by $0.0580$--$0.0829$ on the extra-large variants of YOLOv8, YOLOv10, and YOLO11 (YOLOv8-x, YOLOv10-x, and YOLO11-x) and the extended YOLOv9 variant (YOLOv9-e). 
Because FUSEye requires no camera calibration or dewarping and updates no backbone weights, it fits the label- and training-compute-constrained settings in which robots adapt their detectors; its only deployment overhead is running the same frozen-backbone detector over the five overlapping views. 
In summary, our contributions are listed as follows:

\begin{itemize}
  \item We propose FUSEye, a lightweight modular framework for adapting COCO-pretrained YOLO detectors to fisheye imagery without full-model fine-tuning.

  \item FUSEye consists of three coordinated modules targeting different stages of detection.
  GridViews magnifies compressed peripheral objects, Z-Adapters correct distortion-induced feature misalignment, and AgreeFusion consolidates consistent cross-view detections.

  \item  Experiments on WoodScape verify the effectiveness, efficiency, and transferability of FUSEye. 
  It improves YOLO26-x from $0.1480$ to $0.2661$ $\mathrm{mAP}_{50}$, retains $84.3\%$ of full fine-tuning performance and $97.6\%$ of full-label performance using only $25\%$ of the labels, and consistently improves YOLOv8--11 detectors.
\end{itemize}

\section{Related Work}
\subsection{Fisheye and Omnidirectional Object Detection}
Fisheye object detection research has produced several datasets and benchmarks for studying strong radial distortion, including WoodScape~\cite{woodscape2019}, the OmniCV challenge~\cite{woodscapechallenge2022}, and FishEye8K~\cite{fisheye8k2023}. surveys summarize the main challenges~\cite{fisheyesurvey2023}.
Existing fisheye detectors typically redesign or retrain a complete model. For example, FisheyeDet uses self-study and contour cues~\cite{fisheyedet2020}, Swin-Fisheye adapts Swin transformers~\cite{swinfisheye2024}, FisheyeDetNet targets real-time surround-view systems~\cite{fisheyedetnet2024}, and Rashed~\textit{et al.} provide a generalized detection baseline and representation study~\cite{rashed2021generalized}. 
Streaming and parking settings additionally require efficient inference~\cite{streamingfisheye2023}. 
In parallel, panoramic 360{\textdegree} detection has explored distortion-aware tokenization for overhead fisheye person detection~\cite{panodistorttoken2025}. 
RT-DETR and the YOLO family provide strong COCO-pretrained real-time baselines~\cite{yolov9,yolov10,yolo11,yolo26,rtdetr2023}. 
Most of these methods assume enough fisheye labels and compute to retrain the detector, whereas our goal is to reuse a COCO-pretrained YOLO26 detector with minimal additional training for fisheye imagery.

\subsection{Distortion-Aware Convolutions and Rectification}
Explicit distortion models have been used to reduce fisheye distortion. Rectification networks such as FishFormer~\cite{fishformer2022} and FishRecGAN~\cite{fishrecgan2023} learn to recover perspective images, but require calibration or introduce a separate preprocessing stage. Networks can also be designed to account for distortion. Convolutional filters that are aware of distortion adapt kernels to the panoramic structure~\cite{distortionconv2018}, convolution kernels can be adapted to calibrated fisheye projections~\cite{convkerneladapt2024}, and adaptable deformable convolutions improve fisheye semantic segmentation~\cite{adaptdeform2021,deformfisheye2024}. 
These methods integrate distortion adaptation into the architecture, but they typically retrain the full model and do not preserve the pretrained detector's original behavior. 
Z-Adapters share this goal in a calibration-free, parameter-efficient way. A residual adapter corrects the feature misalignment while a zero-initialized output projection keeps the COCO-pretrained YOLO26 unchanged at the first step of adaptation to the fisheye domain.

\subsection{Parameter-Efficient Fine-Tuning and Visual Adapters}
Parameter-efficient transfer learning has shown that strong performance can be retained while updating only a small fraction of weights. Early residual adapters~\cite{houlsby2019,rebuffi2017}, Low-Rank Adaptation (LoRA)~\cite{lora2022}, and zero-initialized adapter mechanisms~\cite{llamaadapter2023} established the basic recipe, and recent surveys benchmark these methods across vision models~\cite{pettvision2024}. 
Detector-specific Parameter-Efficient Fine-Tuning is also emerging, adapting frozen detection models with lightweight modules~\cite{spotpatch2020,vfmadapter2025,yoloadapter2026}. However, most generic adapters are not distortion-specific. Our Z-Adapters learn residual feature corrections from fisheye data while keeping the COCO-pretrained backbone frozen throughout the entire detector adaptation process.

\subsection{Multi-View Inference and Detection-Level Fusion}
When a single view cannot reliably recognize small or compressed objects, slicing and multi-crop inference can help. Slicing Aided Hyper Inference (SAHI) detects objects in resized overlapping crops and maps the predictions back to the original image coordinates for duplicate suppression~\cite{sahi2022}. AutoFocus selects regions likely to contain small objects for inference at finer scales~\cite{autofocus2019}.
At the detection level, Soft-NMS~\cite{softnms2017} reduces the scores of overlapping boxes, while Weighted Boxes Fusion (WBF)~\cite{wbf2021} merges boxes using confidence-weighted coordinates. Learned NMS learns to suppress duplicate detections from boxes and their scores~\cite{learnednms2017}.
WBF has also been applied to combine predictions from multiple YOLO detectors for fisheye object detection~\cite{tsai2025fisheyefusion}.
In FUSEye, GridViews generates calibration-free overlapping views, and AgreeFusion learns per-cluster agreement scores to promote cross-view-supported candidates while preserving the original confidence threshold for isolated detections observed in only one view.

\section{Method}
\begin{figure*}[t]
  \centering
  \includegraphics[width=0.9\textwidth]{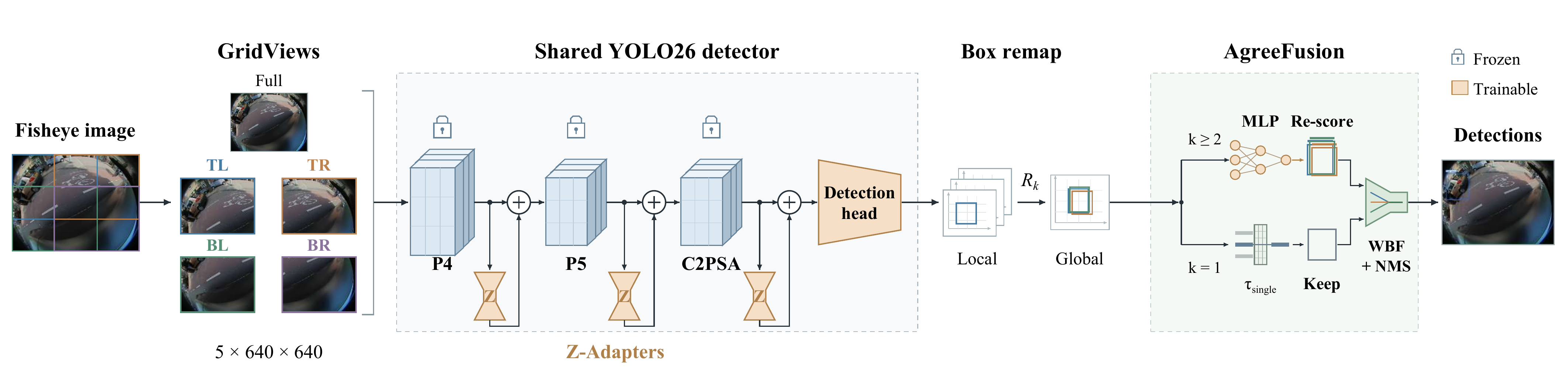}
  \vskip-1ex
  \caption{
    \textbf{Overall architecture of FUSEye.} FUSEye adapts a COCO-pretrained detector to fisheye imagery while keeping its backbone weights frozen. It combines overlapping-view inference, residual feature adaptation, and detection-level agreement fusion. All views share the same adapted detector, and their predictions are remapped to a common coordinate system before fusion. 
    The adapters and detection head are updated during detector adaptation, while a lightweight scorer learns to re-score the resulting multi-view clusters.
  }
  \label{fig:fuseye}
  \vskip-1ex
\end{figure*}

\subsection{Overall Framework}
The detector architecture is shown in Figure~\ref{fig:fuseye}. 
Given a WoodScape fisheye image $x$, GridViews generates $K$ overlapping-grid views (the full image plus four local patches), each of which is resized to $640\times640$ and fed through the same YOLO26 backbone and detection head to produce per-view detections $Y_k$. 
The final AgreeFusion is a detection-level fusion module that fuses the per-view detections for \emph{car}, \emph{person}, and \emph{bus}. 
The pipeline can be summarized as
\begin{equation}
Y = \mathrm{AgreeFusion}\!\left(\bigcup_{k=1}^{K} R_k\!\left(D\!\left(V_k(x)\right)\right)\right),
\label{eq:pipeline}
\end{equation}
where $V_k(x)$ denotes the $k$-th view resized to $640\times640$, $D(\cdot)$ is the shared COCO-pretrained detector (YOLO26 backbone and detection head), $R_k(\cdot)$ remaps the per-view detections back to the full-frame normalized coordinates, and $\mathrm{AgreeFusion}(\cdot)$ clusters the remapped detections by class and IoU, applies agreement-based re-scoring with the learned scorer, fuses consistent boxes, and removes duplicates with Non-Maximum Suppression (NMS). 
All backbone and detection-head weights are initialized from the COCO-pretrained yolo26x.pt checkpoint. The inserted modules are designed so that the initial model behavior is identical to the original detector: the Z-Adapters are exact identity mappings at initialization, GridViews introduces no trainable parameters, and AgreeFusion operates only at the detection level, where a lightweight scorer (about $4.4k$ parameters) is trained separately on cross-view training clusters.
The identity initialization of Z-Adapters ensures that the detector's per-view outputs remain unchanged upon adapter insertion, before any adaptation to the fisheye domain.

\subsection{Overlapping Grid View Generation and Box Remapping}
\begin{figure*}[!t]
  \centering
  \includegraphics[width=0.92\textwidth]{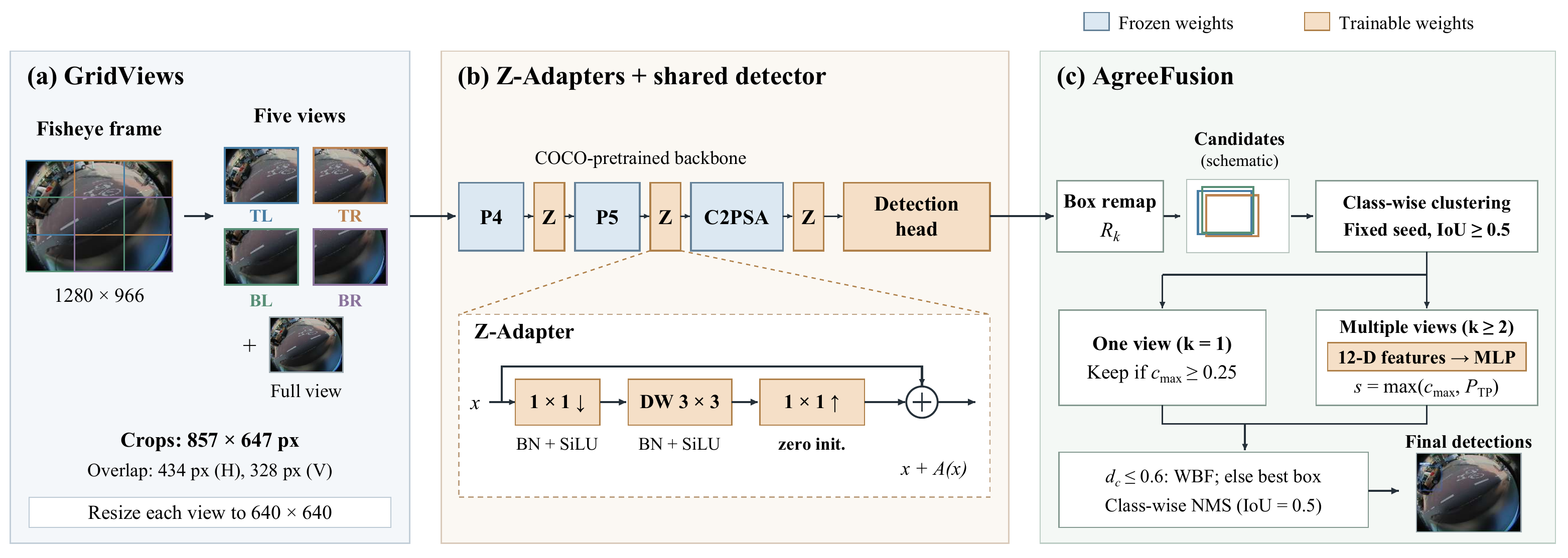}
  \vskip-1.5ex
\caption{Detailed designs of the three FUSEye modules.
(a) GridViews generates the full view and four $857\times647$ corner crops from a $1280\times966$ image. 
Adjacent crops overlap by $423$ pixels horizontally and $319$ pixels vertically. All views are resized to $640\times640$, and their detections are remapped to the original image coordinates.
(b) Zero-initialized residual Z-Adapters are inserted after P4, P5, and C2PSA to model local distortion while preserving the pretrained features.
(c) AgreeFusion clusters aligned boxes by class and IoU and uses learned agreement features to re-score and fuse multi-view detections, followed by class-wise non-maximum suppression.
}
\label{fig:module_details}
\vskip-1ex
\end{figure*}

Objects at the edge of a fisheye image are severely compressed, making them difficult for a COCO-pretrained detector to recognize from the full image alone. As illustrated in Figure~\ref{fig:module_details}(a), GridViews generates $K=5$ views: the full image and four overlapping crops arranged in a $2\times2$ grid. For an original image of size $W_0\times H_0$, each local crop has size $W_c\times H_c$, where $W_0/2<W_c<W_0$ and $H_0/2<H_c<H_0$. The four crops are anchored at the top-left, top-right, bottom-left, and bottom-right corners. Horizontally and vertically adjacent crops overlap by $2W_c-W_0$ and $2H_c-H_0$, respectively. Each view is resized to the network input size $p_w\times p_h$. The full-image view preserves global context, while the local views increase the input-scale size of small objects. The overlapping regions provide opportunities to observe the same object across multiple views, enabling AgreeFusion to assess cross-view agreement after coordinate remapping. Let $(x_1, y_1, x_2, y_2)$ denote the crop region of a view in the original image, with all coordinates measured in pixels; the full-image view corresponds to $(0, 0, W_0, H_0)$.

The remapping of a detection box from a view back to the original fisheye image is written as
\begin{equation}
s_x = \frac{p_w}{x_2 - x_1}, \qquad
s_y = \frac{p_h}{y_2 - y_1},
\end{equation}
\begin{align}
x_{\mathrm{orig}} &= x_1 + x_c / s_x \nonumber \\
y_{\mathrm{orig}} &= y_1 + y_c / s_y \nonumber \\
w_{\mathrm{orig}} &= w / s_x, \quad h_{\mathrm{orig}} = h / s_y.
\end{align}
Here $(x_c, y_c, w, h)$ is the detection box in the view measured in pixels (the view is resized to $p_w \times p_h = 640\times640$), and $(x_1, y_1, x_2, y_2)$ is the crop region in the original image. The remapped boxes are then normalized to the $640\times640$ evaluation space for cross-view fusion:
\begin{align}
\hat{x} &= x_{\mathrm{orig}} / W_0, \quad \hat{w} = w_{\mathrm{orig}} / W_0
    \nonumber \\
\hat{y} &= y_{\mathrm{orig}} / H_0, \quad \hat{h} = h_{\mathrm{orig}} / H_0.
\end{align}
For the full-image view, the remapping reduces to the identity. 
Per-view detections are collected with a confidence threshold of $0.02$ so that low-confidence true positives remain available for the later agreement fusion.
After remapping, boxes whose centers fall outside the full-frame bounds are discarded. 
All remapped boxes are collected as candidate detections for AgreeFusion; class-wise NMS is applied only after cross-view fusion. 
All five views are processed by the same detector, so GridViews costs about five single-view forward passes at $640\times640$. 
This module requires no camera calibration, no image dewarping, and no trainable parameters; it operates purely at the data and coordinate level.
Because the experiments only compare \emph{car}, \emph{person}, and \emph{bus}, we define the following class mapping: COCO person (0) to person, COCO car (2) to car, and COCO bus (5) to bus. 
All other COCO classes are discarded during evaluation to match the target class set.

\subsection{Zero-Initialized Residual Adapter}
Fisheye images exhibit strong appearance variation between the image center and the periphery, and boundary compression makes it difficult for a frozen COCO-pretrained backbone to produce features that are equally reliable across the whole frame. 
Although local crops enlarge small objects at the detector input scale, they do not remove fisheye distortion. 
We therefore introduce lightweight residual adapters to learn feature corrections for the fisheye domain while keeping the pretrained backbone weights frozen. A zero-initialized output projection ensures that each adapter-augmented block produces the same output as its pretrained counterpart at initialization.
In the following, $\mathrm{BN}$ denotes a batch-normalization layer, $\mathrm{DWConv}_{3\times3}$ is a depthwise convolution, and $\sigma$ is the SiLU activation. 
The adapter is written as
\begin{equation}
y = x + A(x),
\end{equation}
where the residual branch $A(x)$ first reduces the channel dimension with a $1\times1$ convolution, applies batch normalization and SiLU, performs local spatial modeling with a $3\times3$ depthwise convolution, applies batch normalization and SiLU again, and finally restores the original channel dimension with a second $1\times1$ convolution. 
The depthwise convolution provides local spatial modeling with only $9C/r$ parameters, much cheaper than a dense $3\times3$ convolution.
Both batch-normalization layers use $\mathrm{affine=False}$, so they add no parameters while stabilizing the internal residual scale. 
Only the output $1\times1$ projection is zero-initialized. 
Therefore, $A(x)=0$ at initialization and the adapter is an identity mapping, leaving the pretrained detector's output unchanged at the first step. Unlike fully zero-initializing every layer, the output projection can receive non-zero gradients at initialization despite its zero-initialized weights, allowing the residual branch to begin learning; after the first optimizer step, gradients flow back to the inner convolutions, and the whole adapter learns a residual correction for the fisheye domain while keeping the pretrained backbone frozen.

We insert the adapter after the P4, P5, and C2PSA stages of the YOLO26 backbone, which in YOLO26x correspond to sequence indices $6$, $8$, and $10$, each with $768$ input channels, where high-level semantic and localization information is most critical for detecting distorted peripheral objects. With a fixed channel reduction ratio of $r=16$, each adapter has $O(C^2/r + 9C/r)$ parameters. 
On YOLO26x, the three adapters introduce $222,480$ parameters in total, while the fine-tuned detection head contains $6,440,952$ parameters, both negligible compared with the frozen backbone.

\subsection{Cross-Projection Agreement Fusion}

A detection can be weak in an individual view yet receive consistent support from other overlapping views. AgreeFusion therefore retains the baseline confidence threshold for isolated detections and uses a learned cluster score to promote supported detections before box consolidation. An early feature-level fusion gate degraded detection quality and was excluded from the final design.

For each class, remapped detections from all views are pooled and sorted by descending confidence. The highest-confidence unassigned detection seeds a cluster containing all remaining unassigned detections with an IoU of at least $0.5$ with the fixed seed box. This process repeats until all detections are assigned. Multiple detections from the same view may enter a cluster, but cross-view support counts only distinct views to avoid counting duplicates.

Each cluster $C$ is represented by a $12$-dimensional feature vector: the matching rate $k/K$, the maximum, mean, and minimum per-view confidence, the noisy-OR score~\cite{viola2005milboost}, the mean pairwise IoU $\mathrm{mIoU}$, the normalized center displacement $d_c$, the normalized box width and height, and a one-hot class indicator. Here, $k$ and $K$ denote the number of distinct matched views and the total number of views, respectively. The class indicator allows the scorer to learn class-specific acceptance policies for different object categories.

The cluster agreement statistics are defined as
\begin{equation}
\mathrm{mIoU}(C)
=\frac{2}{|C|(|C|-1)}
\sum_{i<j}\mathrm{IoU}(b_i,b_j),
\qquad |C|>1,
\label{eq:miou}
\end{equation}
\begin{equation}
d_c(C)
=\frac{\sqrt{\frac{1}{|C|}\sum_i
\lVert\mathbf{c}_i-\bar{\mathbf{c}}\rVert^2}}
{\max(\bar{w},\bar{h},\epsilon)},
\label{eq:center_disp}
\end{equation}
where $b_i=(x_i,y_i,w_i,h_i)$ and $\mathbf{c}_i=(x_i,y_i)$ denote the normalized box and center of detection $i$, and $\bar{\mathbf{c}},\bar{w},\bar{h}$ are the cluster means over all member detections.

Figure~\ref{fig:module_details}(c) uses the cluster-retention rule
\begin{equation}
G(C)=
\begin{cases}
\mathbb{1}[c_{\max}\ge0.25], & k=1,\\
1, & k\ge2.
\end{cases}
\end{equation}
Here, $G(C)=1$ retains the cluster and $0$ discards it; $k$ counts distinct views and $c_{\max}$ is the maximum confidence among all detections within the cluster.
For $k\ge2$, $s(C)=\max(c_{\max},P_{\mathrm{TP}}(C))$, where $s(C)$ and $P_{\mathrm{TP}}(C)$ correspond to the score $s$ and true-positive probability $P_{\mathrm{TP}}$ in the figure.
The scorer is a $12\to96\to32\to1$ MLP with approximately $4.4$k parameters and dropout $0.2$.
It is trained for eight epochs on cross-view clusters using class-weighted binary cross-entropy and pairwise ranking loss, with AdamW, a learning rate of $10^{-3}$, and weight decay of $10^{-4}$.

The box is updated using Weighted Boxes Fusion (WBF) with detection confidences as weights when $d_c\le0.6$; otherwise, the highest-confidence box is kept:
\begin{equation}
\bar{\mathbf{b}}
=\frac{\sum_i\alpha_i\mathbf{b}_i}{\sum_i\alpha_i},
\qquad \alpha_i=c_i^{\mathrm{conf}},
\label{eq:wbf}
\end{equation}
where $\mathbf{b}_i=(x_i,y_i,w_i,h_i)$ and $c_i^{\mathrm{conf}}$ is the detection confidence. Class-wise NMS with an IoU threshold of $0.5$ removes remaining duplicates. This design promotes supported detections with consistent cross-view evidence without lowering the confidence threshold for isolated detections.

\begin{table*}[!t]
  \centering
  \caption{
  Direct-transfer results of COCO-pretrained detectors on WoodScape MVR~\cite{woodscape2019}. Results are reported as percentages. Backbone blocks list the main feature-extraction components; slashes separate component types. Bold and underline denote the best and second-best distinct results, respectively; ties receive the same formatting.}
  \label{tab:direct_transfer}
  \footnotesize
  \setlength{\tabcolsep}{3pt}
  \renewcommand{\arraystretch}{1.06}
  \begin{tabular*}{\textwidth}{@{\extracolsep{\fill}}lclccccc@{}}
    \toprule
    \multicolumn{1}{c}{\multirow{2}{*}{Model}} & \multirow{2}{*}{Variant} & \multicolumn{1}{c}{\multirow{2}{*}{Backbone blocks}} & \multicolumn{2}{c}{Overall} & \multicolumn{3}{c}{Per-class $\mathrm{AP}_{50{:}95}$} \\
    \cmidrule(lr){4-5}\cmidrule(l){6-8}
    & & & $\mathrm{AP}_{50}\uparrow$ & $\mathrm{AP}_{50{:}95}\uparrow$ & Car$\uparrow$ & Person$\uparrow$ & Bus$\uparrow$ \\
    \midrule
    \multirow{5}{*}{YOLOv8~\cite{yolov8}} & Nano & \multirow{5}{*}{C2f} & 5.44 & 3.94 & 6.66 & 5.15 & 0.00 \\
     & Small & & 9.12 & 6.26 & 10.37 & 8.40 & 0.00 \\
     & Medium & & 12.71 & 8.95 & 13.47 & 12.39 & 0.99 \\
     & Large & & 13.07 & 9.66 & 14.25 & 13.73 & 0.99 \\
     & Extra-large & & 12.67 & 9.34 & 14.14 & 13.76 & 0.12 \\
    \midrule
    \multirow{5}{*}{YOLOv9~\cite{yolov9}} & Tiny & \multirow{5}{*}{GELAN} & 5.11 & 3.89 & 6.98 & 4.47 & 0.22 \\
     & Small & & 8.75 & 6.24 & 9.75 & 8.97 & 0.00 \\
     & Medium & & 12.57 & 9.06 & 14.08 & 12.10 & 0.99 \\
     & Compact & & 13.29 & 9.57 & 14.38 & 13.35 & 0.99 \\
     & Extended & & \textbf{14.83} & \underline{10.83} & \underline{15.69} & \underline{15.41} & 1.39 \\
    \midrule
    \multirow{6}{*}{YOLOv10~\cite{yolov10}} & Nano & \multirow{6}{*}{C2f/CIB/PSA} & 5.47 & 3.90 & 6.96 & 4.74 & 0.00 \\
     & Small & & 8.26 & 5.98 & 9.98 & 7.97 & 0.00 \\
     & Medium & & 11.07 & 8.10 & 12.65 & 10.65 & 0.99 \\
     & Balanced & & 12.15 & 8.91 & 13.76 & 12.19 & 0.79 \\
     & Large & & 11.52 & 8.51 & 13.67 & 11.59 & 0.26 \\
     & Extra-large & & 12.82 & 9.57 & 14.86 & 13.37 & 0.50 \\
    \midrule
    \multirow{5}{*}{YOLO11~\cite{yolo11}} & Nano & \multirow{5}{*}{C3k2/C2PSA} & 5.87 & 4.16 & 7.56 & 4.93 & 0.00 \\
     & Small & & 9.55 & 6.69 & 10.67 & 9.28 & 0.12 \\
     & Medium & & 12.18 & 8.86 & 13.52 & 12.07 & 0.99 \\
     & Large & & 13.13 & 9.70 & 15.34 & 12.95 & 0.79 \\
     & Extra-large & & 13.67 & 10.09 & 15.68 & 13.61 & 0.99 \\
    \midrule
    \multirow{5}{*}{YOLO26~\cite{yolo26}} & Nano & \multirow{5}{*}{C3k2/C2PSA} & 5.70 & 4.18 & 7.12 & 5.43 & 0.00 \\
     & Small & & 9.95 & 7.15 & 11.92 & 9.53 & 0.00 \\
     & Medium & & 12.01 & 9.02 & 13.20 & 12.30 & \textbf{1.55} \\
     & Large & & 13.17 & 9.74 & 14.78 & 13.63 & 0.81 \\
     & Extra-large & & \underline{14.80} & \textbf{10.96} & \textbf{15.74} & \textbf{15.67} & \underline{1.49} \\
    \bottomrule
  \end{tabular*}
\end{table*}

\section{Experiments}
\label{sec:experiments}

\subsection{Experimental Setup}
\label{sec:exp_setup}

We train on the front (FV), rear (RV), and mirror-left (MVL) views of WoodScape~\cite{woodscape2019}. The mirror-right (MVR) split is reserved for evaluation to assess performance on a held-out view. Removing training images without annotations for the target classes leaves 6,082 images. We refer to evaluating a COCO-pretrained detector without WoodScape training as \emph{direct transfer}. 
We report box average precision for \emph{car}, \emph{person}, and \emph{bus} using the COCO IoU thresholds. $\mathrm{AP}_{50}$ is evaluated at $\mathrm{IoU}=0.50$ and averaged over the three classes. $\mathrm{AP}_{50{:}95}$ is averaged over the three classes and ten IoU thresholds from 0.50 to 0.95 in steps of 0.05. The per-class columns report $\mathrm{AP}_{50{:}95}$ for each class before averaging across classes.
All results are percentages, and differences are reported in percentage points.

All detectors start from official COCO-pretrained weights and operate at $640\times640$ resolution. Unless stated otherwise, FUSEye uses YOLO26-x with a frozen backbone. 
The detection head and Z-Adapters are optimized for eight epochs with Adam and cosine learning-rate decay.
GridViews generates the full image and four corner-anchored crops.
For the $1280\times966$ input images, we use crops of size $857\times647$, covering approximately $67\%$ of each image dimension.
This yields horizontal and vertical overlaps of $434$ and $328$ pixels, respectively. The AgreeFusion scorer is trained with class-weighted binary cross-entropy and a pairwise ranking loss. 
At inference, we set the per-view candidate confidence threshold to $0.02$ to retain low-confidence candidates for cross-view fusion.
We use $\mathrm{IoU}_{m}=0.5$ for clustering and retain the direct-transfer confidence threshold $\tau_{\mathrm{single}}=0.25$ for single-view detections.
The displacement threshold $\mathrm{disp}_{\max}=0.6$ controls box consolidation: clusters exceeding this threshold retain the highest-confidence box instead of averaging potentially inconsistent locations. 
The detection head, adapters, and scorer together contain approximately $6.67$ million trainable parameters, of which approximately $227\,\mathrm{k}$ are introduced by FUSEye.
Full fine-tuning updates all detector parameters for ten epochs using the Ultralytics training recipe and serves as a fully adapted reference.

\subsection{Main Results}
\label{sec:main_results}

Table~\ref{tab:direct_transfer} shows that scaling helps primarily at the lower end of each family. Gains diminish for larger variants and are sometimes non-monotonic, with the largest variants of all five families remaining within a narrow $12.67$--$14.83$ $\mathrm{AP}_{50}$ range. Full fine-tuning, by contrast, raises YOLO26-x from $14.80$ to $31.56$.
Within the tested models, target-domain adaptation is therefore more consequential than further increasing the capacity of a COCO-pretrained detector. The class breakdown reveals a limitation shared across families: \emph{bus} detection remains weak compared with \emph{car} and \emph{person}. 
We select YOLO26-x for subsequent experiments because it leads $\mathrm{AP}_{50{:}95}$, \emph{car}, and \emph{person}, while its $\mathrm{AP}_{50}$ is only $0.03$ points below the best result across all families.

\begin{table}[!t]
  \centering
  \caption{Comparison of adaptation strategies for YOLO26-x on WoodScape MVR.
  Bold and underline denote the best and second-best results, respectively.}
  \label{tab:main_results}
  \footnotesize
  \setlength{\tabcolsep}{1pt}
  \renewcommand{\arraystretch}{1.12}
  \begin{tabular*}{\columnwidth}{@{\extracolsep{\fill}}lcccccc@{}}
    \toprule
    \multicolumn{1}{c}{\multirow{2}{*}{Method}} & \multirow{2}{*}{\shortstack{Backbone\\frozen}} & \multicolumn{2}{c}{Overall} & \multicolumn{3}{c}{Per-class $\mathrm{AP}_{50{:}95}$} \\
    \cmidrule(lr){3-4}\cmidrule(l){5-7}
    & & $\mathrm{AP}_{50}\uparrow$ & $\mathrm{AP}_{50{:}95}\uparrow$ & Car$\uparrow$ & Person$\uparrow$ & Bus$\uparrow$ \\
    \midrule
    Direct transfer & Yes & 14.80 & 10.96 & 15.74 & 15.67 & \underline{1.49} \\
    \rowcolor{black!4}
    FUSEye & Yes & \underline{26.61} & \underline{17.26} & \underline{26.83} & \textbf{24.06} & 0.87 \\
    Full fine-tuning & No & \textbf{31.56} & \textbf{22.82} & \textbf{42.48} & \underline{22.44} & \textbf{3.54} \\
    \bottomrule
  \end{tabular*}
\end{table}

Table~\ref{tab:main_results} compares direct transfer, FUSEye, and full fine-tuning on YOLO26-x. With the backbone frozen, FUSEye improves $\mathrm{AP}_{50}$ by $11.81$ points and $\mathrm{AP}_{50{:}95}$ by $6.30$ points, recovering $70.5\%$ and $53.1\%$ of the corresponding gains from full fine-tuning.
The lower recovery under $\mathrm{AP}_{50{:}95}$ indicates a larger relative gap at stricter localization thresholds. FUSEye outperforms full fine-tuning on \emph{person} but remains substantially lower on \emph{car}, which accounts for the largest remaining gap. 
\emph{Bus} remains a separate limitation: FUSEye falls below direct transfer, whereas full fine-tuning improves its performance on this challenging object category.

\subsection{Ablation Study}
\label{sec:ablation}

\begin{table}[!tb]
  \centering
  \caption{Component ablation on YOLO26-x. GV, ZA, and AF denote GridViews, Z-Adapters, and AgreeFusion, respectively; the first row disables all three components. Bold and underline denote the best and second-best distinct results, respectively.}
  \label{tab:ablation}
  \footnotesize
  \setlength{\tabcolsep}{1pt}
  \renewcommand{\arraystretch}{1.12}
  \begin{tabular*}{\columnwidth}{@{\extracolsep{\fill}}cccccccc@{}}
    \toprule
    \multirow{2}{*}{GV} & \multirow{2}{*}{ZA} & \multirow{2}{*}{AF} & \multicolumn{2}{c}{Overall} & \multicolumn{3}{c}{Per-class $\mathrm{AP}_{50{:}95}$} \\
    \cmidrule(lr){4-5}\cmidrule(l){6-8}
    & & & $\mathrm{AP}_{50}\uparrow$ & $\mathrm{AP}_{50{:}95}\uparrow$ & Car$\uparrow$ & Person$\uparrow$ & Bus$\uparrow$ \\
    \midrule
    -- & -- & -- & 14.62 & 11.05 & 15.63 & 16.27 & \underline{1.25} \\
    \checkmark & -- & -- & 15.49 & 11.22 & 13.60 & 18.51 & \textbf{1.55} \\
    -- & \checkmark & -- & 17.75 & 13.01 & \underline{20.37} & 17.76 & 0.89 \\
    \checkmark & \checkmark & -- & \underline{20.39} & \underline{14.03} & 19.74 & \underline{21.45} & 0.89 \\
    \checkmark & -- & \checkmark & 18.91 & 12.95 & 17.36 & 20.43 & 1.06 \\
    -- & \checkmark & \checkmark & 17.75 & 13.01 & \underline{20.37} & 17.76 & 0.89 \\
    \rowcolor{black!4}
    \checkmark & \checkmark & \checkmark & \textbf{26.61} & \textbf{17.26} & \textbf{26.83} & \textbf{24.06} & 0.87 \\
    \bottomrule
  \end{tabular*}
\end{table}

Table~\ref{tab:ablation} shows complementary effects from GridViews and Z-Adapters.
GridViews alone changes overall $\mathrm{AP}_{50{:}95}$ only marginally, but this aggregate masks a gain on \emph{person} and a loss on \emph{car}. Z-Adapters yield a larger overall gain, driven mainly by improved \emph{car} detection.
Combining the two produces the best \emph{person} score without fusion, but leaves \emph{car} slightly below Z-Adapters alone.
Enlarging local content and adapting features therefore address different class-wise errors, while their combination still leaves a trade-off between \emph{person} and \emph{car} performance.

The contribution of AgreeFusion depends on the candidates it receives. Added to GridViews, it improves $\mathrm{AP}_{50}$ and reverses the \emph{car} loss relative to the baseline. Added after both GridViews and Z-Adapters, it produces larger gains in both overall metrics, improving \emph{car} and \emph{person}.
This conditional improvement is consistent with fusion benefiting from the adapted candidate set, although the table does not isolate the candidate-level mechanism.
Without GridViews, adding AgreeFusion to Z-Adapters does not change the result because no cross-view candidates are available.
The score on \emph{Bus} does not follow the \emph{car} and \emph{person}'s trend: configurations containing Z-Adapters consistently score lower than those without them.
Given the near-zero scores and the absence of uncertainty estimates, this pattern is treated as a recurring limitation rather than a causal effect.

\subsection{Training-Set Size}
\label{sec:training_size}

\begin{table}[!t]
  \centering
  \caption{Results with $25\%$ and $100\%$ of the training set. Each $25\%$ subset contains $1,520$ images; mean $\pm$ sample standard deviation is computed over three subsets.}
  \label{tab:label_efficiency}
  \footnotesize
  \setlength{\tabcolsep}{1pt}
  \renewcommand{\arraystretch}{1.16}
  \begin{tabular*}{\columnwidth}{@{\extracolsep{\fill}}lccccc@{}}
    \toprule
    \multicolumn{1}{c}{\multirow{2}{*}{Metric}} & \multicolumn{4}{c}{25\% training set} & 100\% \\
    \cmidrule(lr){2-5}\cmidrule(l){6-6}
    & Subset 1 & Subset 2 & Subset 3 & Mean $\pm$ SD & Full \\
    \midrule
    $\mathrm{AP}_{50}\uparrow$ & 26.21 & 26.12 & 25.59 & \cellcolor{black!4}$25.97\pm0.34$ & 26.61 \\
    $\mathrm{AP}_{50{:}95}\uparrow$ & 17.04 & 16.91 & 16.56 & \cellcolor{black!4}$16.84\pm0.25$ & 17.26 \\
    \addlinespace[1pt]
    $\mathrm{AP}^{\mathrm{car}}_{50{:}95}\uparrow$ & 26.21 & 26.19 & 26.11 & \cellcolor{black!4}$26.17\pm0.05$ & 26.83 \\
    $\mathrm{AP}^{\mathrm{person}}_{50{:}95}\uparrow$ & 23.10 & 23.04 & 22.24 & \cellcolor{black!4}$22.79\pm0.48$ & 24.06 \\
    $\mathrm{AP}^{\mathrm{bus}}_{50{:}95}\uparrow$ & 1.82 & 1.48 & 1.33 & \cellcolor{black!4}$1.55\pm0.25$ & 0.87 \\
    \bottomrule
  \end{tabular*}
\end{table}

We repeat training on three independently sampled subsets of $1,520$ images, using the same subset for the adapters, detection head, and fusion scorer.
As shown in Table~\ref{tab:label_efficiency}, the $25\%$ subsets retain $97.6\%$ of the full-data result on both overall metrics on average.
Every subset remains slightly below the full-data result, but the absence of variability estimates for full-data training prevents a significance claim.
The class-wise behavior is less uniform: \emph{car} shows less variability across subsets and a smaller deficit than \emph{person}. 
Within this comparison, the observed gain from additional labels is larger for \emph{person}.
The score on \emph{Bus} is higher in all three subset runs than in the full-data result, but its near-zero scores and the missing full-data variability preclude interpreting this as an advantage of using fewer labels.

\subsection{Cross-Model Evaluation}
\label{sec:other_models}

\begin{table}[!t]
  \centering
  \caption{Cross-model evaluation on WoodScape MVR using the protocol in Sec.~\ref{sec:exp_setup}.
  The evaluated variants are Extra-large for YOLOv8, YOLOv10, and YOLO11, and Extended for YOLOv9.
  Base denotes COCO-pretrained direct transfer.
  Results are percentages; bold marks the better result in each pair.}
  \label{tab:transfer}
  \footnotesize
  \setlength{\tabcolsep}{0.8pt}
  \renewcommand{\arraystretch}{1.10}
  \begin{tabular*}{\columnwidth}{@{\extracolsep{\fill}}llccccc@{}}
    \toprule
    \multicolumn{1}{c}{\multirow{2}{*}{Model}} & \multicolumn{1}{c}{\multirow{2}{*}{Method}} & \multicolumn{2}{c}{Overall} & \multicolumn{3}{c}{Per-class $\mathrm{AP}_{50{:}95}$} \\
    \cmidrule(lr){3-4}\cmidrule(l){5-7}
    & & $\mathrm{AP}_{50}\uparrow$ & $\mathrm{AP}_{50{:}95}\uparrow$ & Car$\uparrow$ & Person$\uparrow$ & Bus$\uparrow$ \\
    \midrule
    \multirow{2}{*}{YOLOv8-x~\cite{yolov8}} & Base & 11.89 & 9.05 & 12.79 & 13.36 & \textbf{0.99} \\
      & FUSEye & \textbf{17.69} & \textbf{11.50} & \textbf{17.70} & \textbf{16.35} & 0.45 \\
    \midrule
    \multirow{2}{*}{YOLOv9-e~\cite{yolov9}} & Base & 13.84 & 10.23 & 14.11 & 15.43 & \textbf{1.16} \\
      & FUSEye & \textbf{22.13} & \textbf{13.89} & \textbf{19.95} & \textbf{20.90} & 0.83 \\
    \midrule
    \multirow{2}{*}{YOLOv10-x~\cite{yolov10}} & Base & 12.25 & 9.16 & 13.40 & 13.09 & \textbf{0.99} \\
      & FUSEye & \textbf{19.00} & \textbf{12.39} & \textbf{18.28} & \textbf{18.64} & 0.24 \\
    \midrule
    \multirow{2}{*}{YOLO11-x~\cite{yolo11}} & Base & 12.49 & 9.38 & 13.98 & 13.16 & \textbf{0.99} \\
      & FUSEye & \textbf{19.70} & \textbf{12.48} & \textbf{18.78} & \textbf{17.92} & 0.74 \\
    \bottomrule
  \end{tabular*}
\end{table}

Table~\ref{tab:transfer} presents the evaluation of generalization capacity across detectors.
FUSEye improves $\mathrm{AP}_{50}$ by $5.80$--$8.29$ points and $\mathrm{AP}_{50{:}95}$ by $2.45$--$3.66$ points for every additional detector. 
The \emph{car} gains are relatively stable across the four families at $4.80$--$5.84$ points, whereas person gains vary from $2.99$ to $5.55$ points. 
The consistent improvement across detectors using C2f, GELAN, CIB/PSA, and C3k2/C2PSA backbone blocks indicates that the benefit is not confined to the YOLO26-x architecture.
It is not class-uniform, however: \emph{bus} decreases by $0.25$--$0.75$ points in every pair, reproducing the limitation observed with YOLO26-x.

\begin{figure}[h]
  \centering
  \includegraphics[width=\columnwidth]{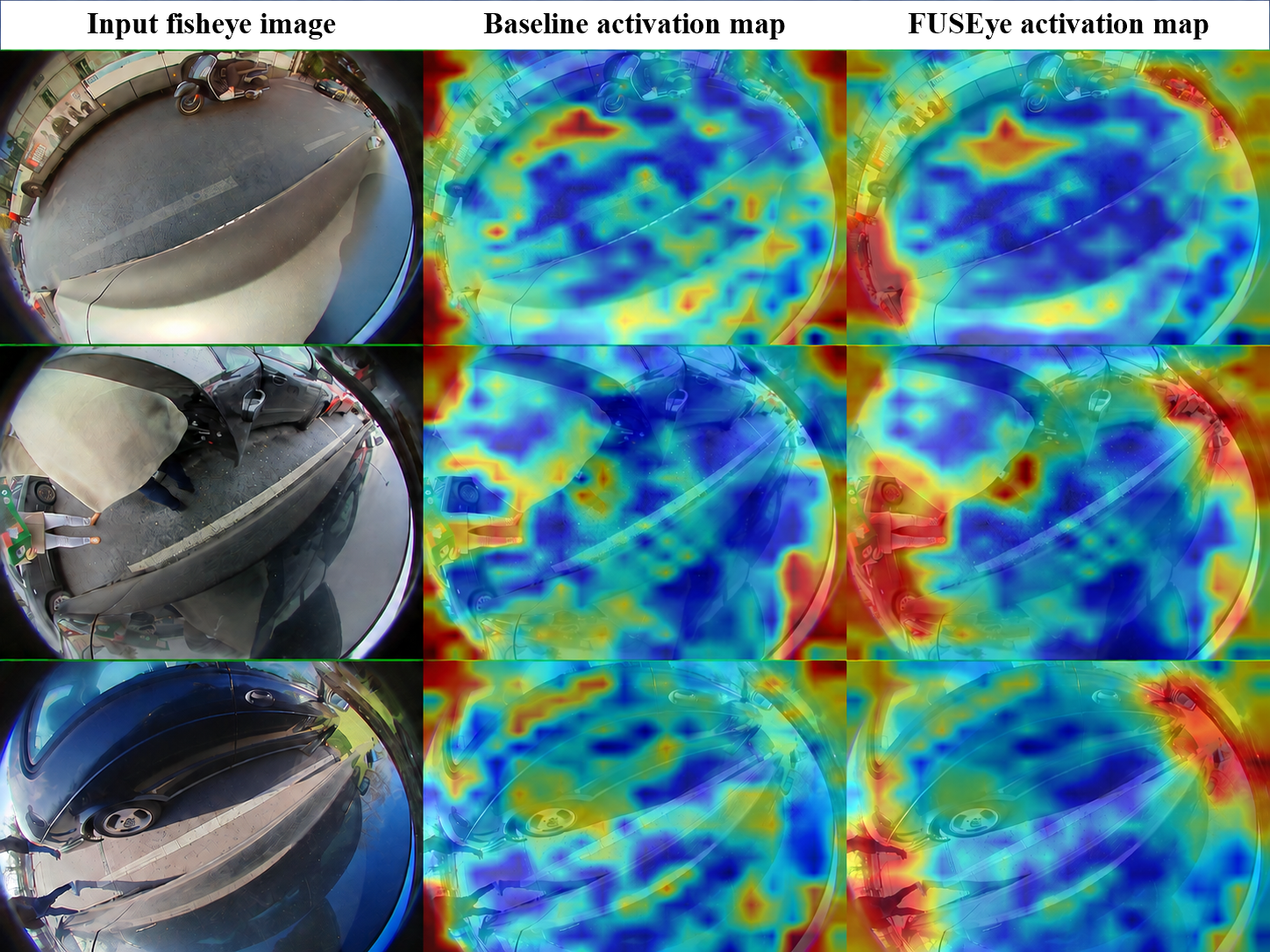}
  \caption{C2PSA activation maps from the YOLO26-x backbone for direct transfer and FUSEye. Warmer colors indicate stronger responses.}
  \label{fig:heatmap}
\end{figure}

\subsection{Qualitative Results}
\label{sec:qualitative}
\begin{figure}[!t]
  \centering
  \includegraphics[width=\columnwidth]{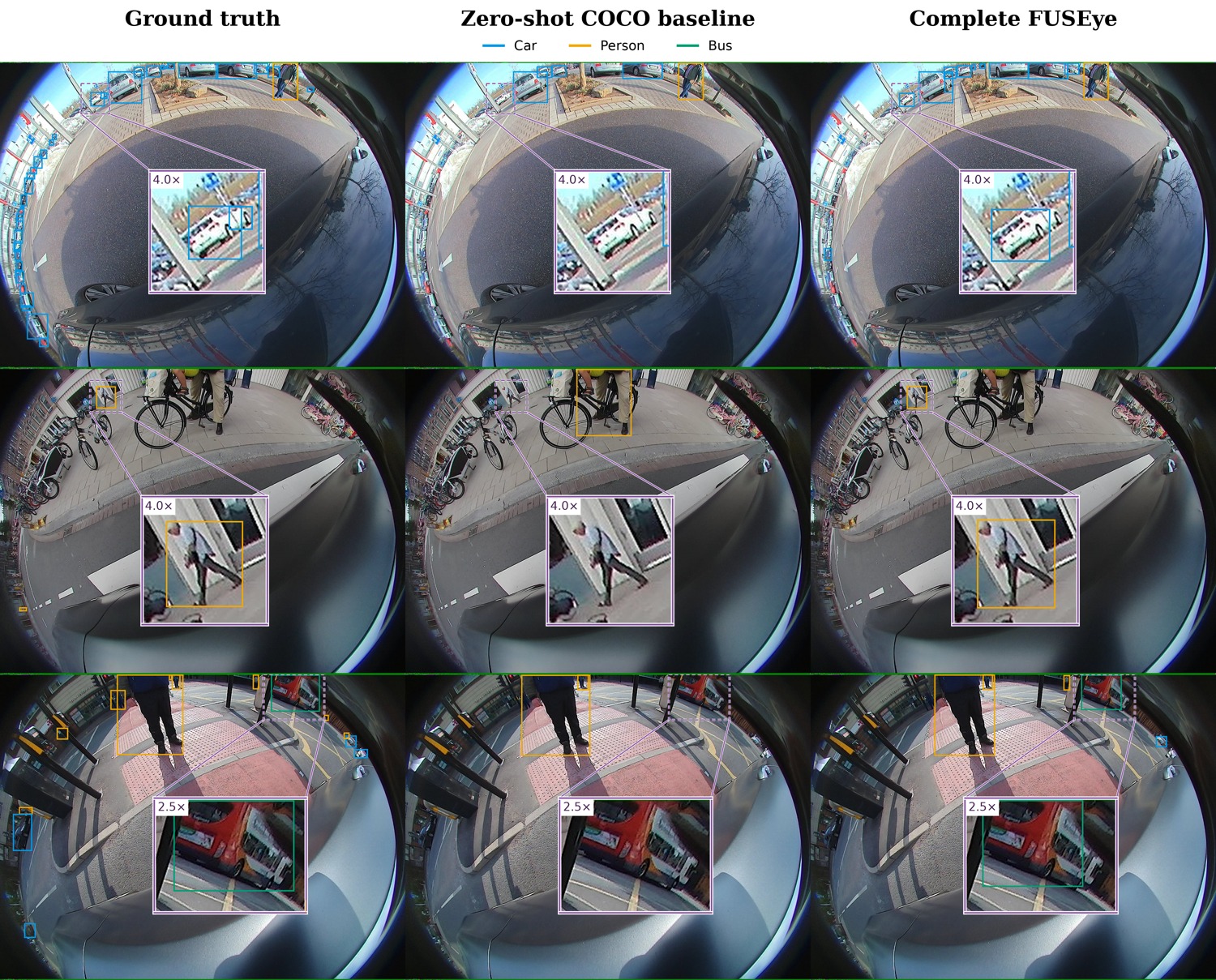}
  \caption{Columns show ground truth, zero-shot COCO baseline, and complete FUSEye; insets enlarge representative detections on WoodScape MVR.}
  \label{fig:detection}
\end{figure}

Figures~\ref{fig:heatmap} and~\ref{fig:detection} show activation and detection changes after adaptation.
In Figure~\ref{fig:detection}, FUSEye recovers peripheral objects missed or weakly detected by the zero-shot baseline, consistent with GridViews enlarging content and Z-Adapters correcting distorted features.
AgreeFusion supports recurring weak detections; false positives and localization errors show agreement does not ensure correctness.

\section{Conclusion}
\label{sec:conclusion}
Scaling COCO-pretrained detectors yields diminishing direct-transfer gains on WoodScape MVR, leaving a clear fisheye transfer gap. FUSEye addresses this gap while keeping the pretrained backbone frozen through overlapping global and local views, zero-initialized residual adapters, and detection-level agreement fusion.
On YOLO26-x, FUSEye raises $\mathrm{AP}_{50}$ from $14.80$ to $26.61$ and $\mathrm{AP}_{50{:}95}$ from $10.96$ to $17.26$, recovering $70.5\%$ and $53.1\%$ of the gains from full fine-tuning. 
It introduces approximately $227\,\mathrm{k}$ additional parameters. 
The Z-Adapters are trained jointly with the existing detection head, while the scoring component within AgreeFusion is trained separately. Using only $25\%$ of the training images, FUSEye retains $97.6\%$ of its full-data $\mathrm{AP}_{50}$ and $\mathrm{AP}_{50{:}95}$. The pipeline improves $\mathrm{AP}_{50}$ by $5.80$--$8.29$ points in four additional YOLO detectors.
Ablation results attribute these gains to complementary effects: GridViews strengthens person detection, Z-Adapters provide a car-led improvement, and AgreeFusion benefits most from adapted multi-view candidates.
We will explore parameter-efficient adaptation to improve detection accuracy while reducing trainable parameters.

\bibliographystyle{IEEEtran}
\bibliography{bib}

\end{document}